\documentclass[journal]{IEEEtran}
\IEEEoverridecommandlockouts                              

\newtheorem{Remark}{Remark}

\newtheorem{Problem}{Problem}
\newenvironment{Proof}{\noindent{\em Proof:\/}}{\hfill $\Box$\par}
\newtheorem{Theorem}{Theorem}

\usepackage[ruled,linesnumbered]{algorithm2e}
\usepackage{verbatim}
\usepackage{graphicx}
\usepackage{subfigure}
\usepackage{amsmath,bm}
\usepackage{amssymb}
\usepackage{algpseudocode}
\usepackage{cite}
\usepackage{hyperref}
\usepackage{booktabs}
\usepackage{multirow}
\usepackage{orcidlink}

\allowdisplaybreaks

\algdef{SE}[DOWHILE]{Do}{doWhile}{\algorithmicdo}[1]{\algorithmicwhile\ #1}%

\newcommand{\mathactivatecomma}{%
  \begingroup\lccode`~=`\,
  \lowercase{\endgroup\edef~}{\mathchar\the\mathcode`\,\penalty0 }}

\algnewcommand{\Initialize}[1]{%
  \State \textbf{Initialize: $i \in \mathcal{V}$}
  \Statex \hspace*{\algorithmicindent}\parbox[t]{.8\linewidth}{\raggedright #1}
}

\algnewcommand{\Iteration}[1]{%
  \State \textbf{Iteration $(k\geq 0)$: $i \in \mathcal{V}$}
  \Statex \hspace*{\algorithmicindent}\parbox[t]{.8\linewidth}{\raggedright #1}
}

\algnewcommand{\Output}[1]{%
  \State \textbf{Output: $i \in \mathcal{V}$}
  \Statex \hspace*{\algorithmicindent}\parbox[t]{.8\linewidth}{\raggedright #1}
}

\ifCLASSINFOpdf
\else
\fi
\begin{document}
%
\title{Low-Complexity Cooperative Payload Transportation for Nonholonomic Mobile Robots Under Scalable Constraints}
%
%
%

\author{Renhe~Guan~\orcidlink{0000-0001-5961-4531}, Yuanzhe~Wang~\orcidlink{https://orcid.org/0000-0002-1683-4849}, Tao~Liu~\orcidlink{0000-0001-8068-8025}
        and~Yan~Wang~\orcidlink{0000-0002-8644-8791}
\thanks{R. Guan and Yan Wang are with the School of Mechanical Electrical Engineering and Automation, Harbin Institute of Technology Shenzhen, Shenzhen, China. (email: guanrenhe142@gmail.com; wang.yan@hit.edu.cn) (Corresponding author: Yan Wang.)

Yuanzhe Wang is with School of Control Science and Engineering, Shandong University, Jinan, Shandong, China.

T. Liu is with the School of Automation and Intelligent Manufacturing, Southern University of Science and Technology, Shenzhen 518055, China. (email: liut6@sustech.edu.cn)}
}

%
%

\markboth{IEEE Transactions on Industrial Cyber-Physical Systems}%
{Shell \MakeLowercase{\textit{et al.}}: Bare Demo of IEEEtran.cls for IEEE Journals}
%



\maketitle

\begin{abstract}
Cooperative transportation, a key aspect of logistics cyber-physical systems (CPS), is typically approached using distributed control and optimization-based methods. The distributed control methods consume less time, but poorly handle and extend to multiple constraints. Instead, optimization-based methods handle constraints effectively, but they are usually centralized, time-consuming and thus not easily scalable to numerous robots. To overcome drawbacks of both, we propose a novel cooperative transportation method for nonholonomic mobile robots by improving conventional formation control, which is distributed, has a low time-complexity and accommodates scalable constraints. The proposed control-based method is testified on a cable-suspended payload and divided into two parts, including robot trajectory generation and trajectory tracking. Unlike most time-consuming trajectory generation methods, ours can generate trajectories with only constant time-complexity, needless of global maps. As for trajectory tracking, our control-based method not only scales easily to multiple constraints as those optimization-based methods, but reduces their time-complexity from polynomial to linear. Simulations and experiments can verify the feasibility of our method.         
\end{abstract}

%
\IEEEpeerreviewmaketitle
\section{Introduction}
\IEEEPARstart{R}{ecently}, logistics cyber-physical systems (CPS), particularly multi-robot cooperative transportation, have garnered increasing attention due to their advantages, such as cost reduction and enhanced productivity \cite{G.A.2009, B.L.2023, Y.L.2023, J.F.2011, W.Z.2016, P.G.2004, J.F.2008,W.W.2020,Z.W.2016, J.A.2015, J.A2.2015, D.K.2021, N.L.2020, B.E.2020, D.M.2011, F.A.2012, Y.C.2015}. In this scenario, robots are required to coordinately transport the payload from a starting place to the desired destination quickly. Typically, the robot formation is subject to numerous constraints in practical transportation, such as obstacle avoidance, inter-robot collision avoidance, velocity constraints, payload protection, nonholonomic kinematics, etc. So far, how to overcome as many constraints as possible in the shortest time has become an important issue in cooperative transportation problems. 

Most cooperative transportation algorithms are based on two frameworks, including distributed control \cite{J.F.2011, W.Z.2016, P.G.2004, J.F.2008,W.W.2020,Z.W.2016} and optimization \cite{J.A.2015, J.A2.2015, D.K.2021, N.L.2020, B.E.2020, D.M.2011, F.A.2012, Y.C.2015}. However, neither of both can accommodate scalable constraints with relatively low time-complexity. Specifically, distributed control methods \cite{J.F.2011, W.Z.2016, P.G.2004, J.F.2008,W.W.2020,Z.W.2016} scale better in the number of robots and consume less time, but do not scale easily to multiple constraints. The controllers in these papers are implemented by consuming at most linear time $O(n)$, where $n$ is the number of robots. For example, papers like \cite{W.Z.2016, Z.W.2016} adopt a pushing strategy to deliver the payload to its destination, requiring only constant time to estimate the velocity of the payload and implement the designed controller. Other papers including \cite{J.F.2011, P.G.2004, J.F.2008, W.W.2020} use a caging strategy, which requires the robot to receive velocity or position information from its neighboring robots for controller design, consuming time roughly linearly proportional to the number of robots. Besides, due to the low time consumption of controller implementation, these algorithms can easily be extended to numerous robots. For instance, authors in \cite{W.Z.2016, Z.W.2016} perform simulations with over 20 robots, while papers including \cite{J.F.2008} use 8 to 20 robots to complete transportation experiments in real time.

However, distributed control methods are weak in enforcing multiple constraints as mentioned in \cite{B.E.2020}. They are usually designed for holonomic robots with specific constraints and can not be extended to new constraints by simple modifications. Although some control methods, such as null-space approach \cite{G.A.2009}, can handle constraints in a hierarchy of priorities, they only accommodate a limited number and type of constraints, excluding velocity constraints. In contrast, optimization methods can overcome it. Papers like \cite{B.E.2020} model cooperative transportation as a multi-robot trajectory generation problem with multiple constraints and use optimization methods to solve it. Then, the transportation task is completed by robots tracking the generated trajectories. Such problem can also be solved by centralized mixed integer programming \cite{D.M.2011} and sequential convex programming \cite{F.A.2012, Y.C.2015}. However, these methods \cite{B.E.2020, D.M.2011, F.A.2012, Y.C.2015} require environment maps for global planning and consume much time, where the computation cost of \cite{F.A.2012} grows exponentially with the number of robots or constraints. Thus, they are incapable of working online and adapting to environments without prior maps.

To tackle it, some papers including \cite{J.A.2015, J.A2.2015, D.K.2021, N.L.2020} improve optimization methods to enable transportation systems to make decisions online.
Specifically, authors in \cite{J.A.2015, J.A2.2015} formulate the transportation problem as a convex optimization problem and solve it in a distributed fashion to speed up. However, these methods are still time-consuming, with each global trajectory update taking several seconds and local velocity update taking several hundred milliseconds for only three robots. In \cite{D.K.2021}, the transportation problem is solved using recently popular hierarchical
quadratic programming method (HQP) proposed by \cite{E.A.2014}. Nevertheless, HQP and similar cascade quadratic programming method in \cite{O.K.2011} are centralized, where the number of constraints and the time consumed increase polynomially with the number of robots. Furthermore, the decentralized leader-follower model predictive control method is applied to achieve cooperative transportation in \cite{N.L.2020}, where the control frequency is only 10Hz with only two robots. In general, although the time consumed by these algorithms may be acceptable in some specific cases, they are much worse than distributed control methods in terms of time-complexity.

To solve these problems, we design a cooperative transportation method for a cable-suspended payload using nonholonomic mobile robots by combining advantages of low time-complexity in distributed control with the ease of handling constraints in optimization. Our method is developed by improving the conventional distributed formation control, whose main contributions are summarized below. 

\begin{itemize}
    \item Different from most transportation methods based on distributed control \cite{J.F.2011, W.Z.2016, P.G.2004, J.F.2008,W.W.2020,Z.W.2016}, our approach is easily extended to multiple constraints, through a designed multi-process control framework. Besides, velocity constraints and scalable constraints that are not addressed in null-space approach \cite{G.A.2009} are included.  
    \item Unlike most related optimization-based methods with the polynomial and exponential time-complexity \cite{J.A.2015, J.A2.2015, D.K.2021, N.L.2020, B.E.2020, D.M.2011, F.A.2012, Y.C.2015, E.A.2014, O.K.2011}, ours consumes only linear time-complexity in total with the number of robots and constraints.
    \item In contrast to those centralized offline methods relying on prior environment maps \cite{D.M.2011, F.A.2012}, ours is distributed and can react to environments without maps in real time.   
\end{itemize}

The remainder of this paper is structured as follows. Section \uppercase\expandafter{\romannumeral2} presents the constrained cooperative transportation problem and our proposed scheme, which includes trajectory generation and trajectory tracking using a leader-follower structure. Section \uppercase\expandafter{\romannumeral3} details the constant time-complexity trajectory generation methods for both the leader and followers. Section \uppercase\expandafter{\romannumeral4} discusses trajectory tracking, where followers maintain constraints with linear time-complexity. In Section \uppercase\expandafter{\romannumeral5}, numerical simulations and real experiments validate our methods. Finally, Section \uppercase\expandafter{\romannumeral6} concludes the paper.

\section{System and Scheme Overview}\label{sec:problem_formulation}
\subsection{Structure of Cooperative Transportation System}
In this paper, a team of nonholonomic mobile robots are applied to accomplish the cooperative transportation task. For each robot $i$, the kinematics model is represented as
\begin{equation} \label{eq1}
\dot{x}_i = v_i {\rm cos}\theta_i,\;
\dot{y}_i = v_i {\rm sin}\theta_i,\;
\dot{\theta}_i = \omega_i
\end{equation} 
where $\bm{p}_i=[x_i(t),y_i(t)]^T$ is the position, $\theta_i(t) \in (-\pi,\pi]$ is the heading angle of robot $i$ with respect to the global frame $G$. The linear velocity of robot $i$ is $v_i$ and the angular velocity is $\omega_i$, while the velocity vector is $\bm{u}_i=[v_i,\omega_i]^T$. Given that most commercialized mobile robots provide velocity inputs, we do not consider the robot dynamics. 

\begin{figure}[htbp]
\vspace{-0.45cm}
\centering
\includegraphics[width=3.2in]{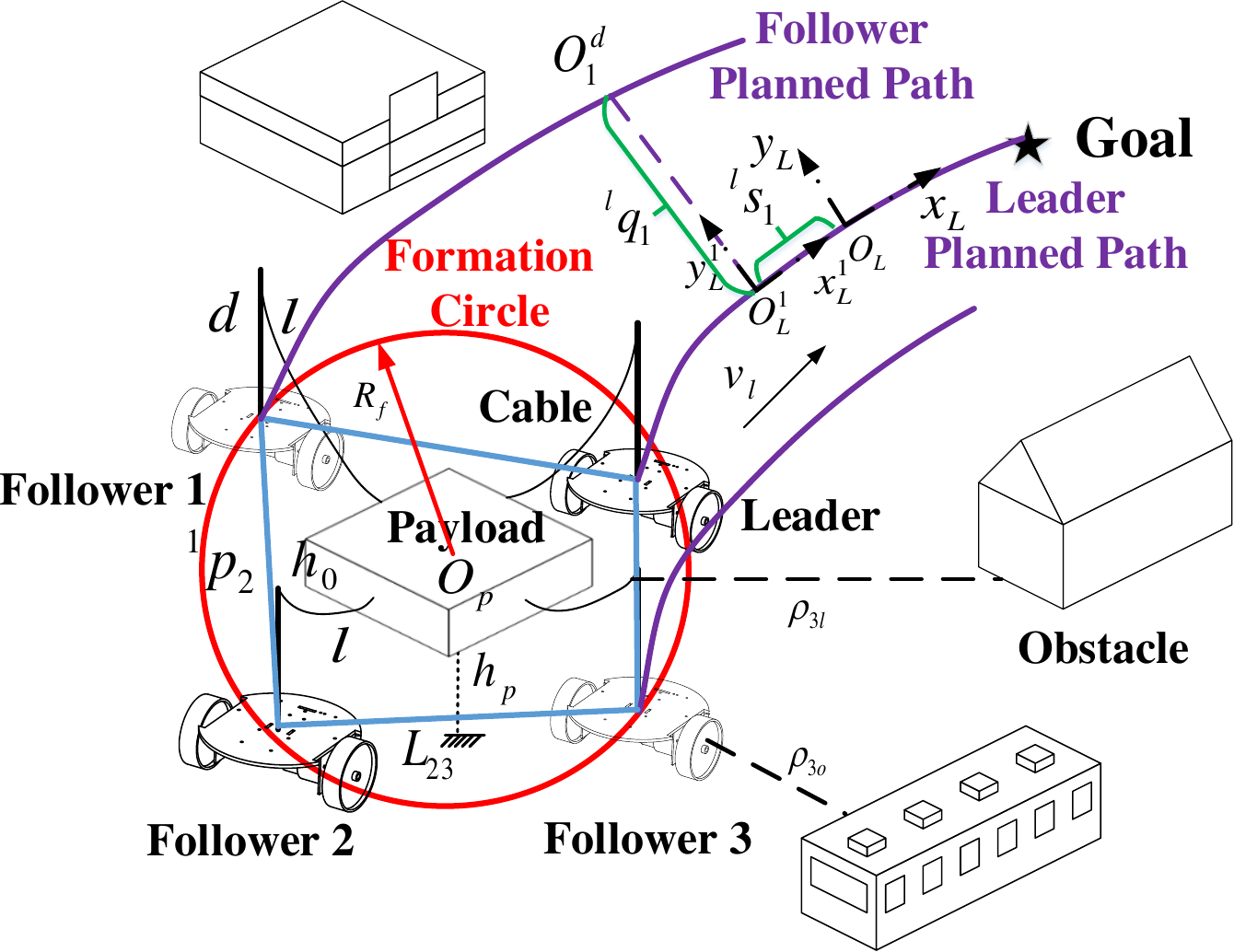}  
\caption{The cooperative transportation system in obstacle environments}
\label{Transport}
\vspace{-0.3cm}
\end{figure}

As in Fig \ref{Transport}, the payload is raised by mobile robots with cables attached to robots at one end and to the payload at the other. Assume that the cable on each robot is fixed at height $d$, the length of cable is $l$ and the goal position is $\bm{p}_g = [x_g,y_g]^T$. The robot in the formation needs to efficiently perform two tasks in a distributed manner, including $1)$ generating a reference trajectory to $\bm{p}_g$ and $2)$ overcoming various constraints during transportation.

For fast trajectory generation, it is better for robots to adopt a control framework rather than a time-consuming trajectory optimization framework. To further increase efficiency, the transportation system can operate in a leader-follower configuration, with one robot in the front as the leader $l$ and the other $n_r$ robots as followers. The leader is only responsible for guiding the whole formation to $\bm{p}_g$ safely, while followers follow the leader and preserve constraints with the leader and other followers. In this way, followers can save time in trajectory generation and focus on overcoming constraints.

Here, all robots form the set $\mathcal{R}$, all followers form $\mathcal{F}$ and the neighboring set of $i$ is $\mathcal{N}_i=\mathcal{R}\backslash \{i\}$. Each follower is equipped with a cheap monocular limited filed-of-view (FOV) camera to detect the leader and a laser scanner to detect obstacles. Without loss of generality, the origin of the global frame $G$ is selected as the initial position of the leader $l$, whose forward direction coincides with $x$ axis. The position of leader is denoted by $\bm{p}_l$, the position of the $k^{\text{th}}\,(k\le n_r)$ follower is $\bm{p}_k$ and the relative position between robots $i$ and $j$ in $i$'s local frame is ${}^i\bm{p}_j$, where $i, j\in \mathcal{R}$. We consider the transportation is roughly completed when the leader arrives $\bm{p}_g$. 

\begin{figure}[htbp]
\vspace{-0.45cm}
\centering
\includegraphics[width=3.1in]{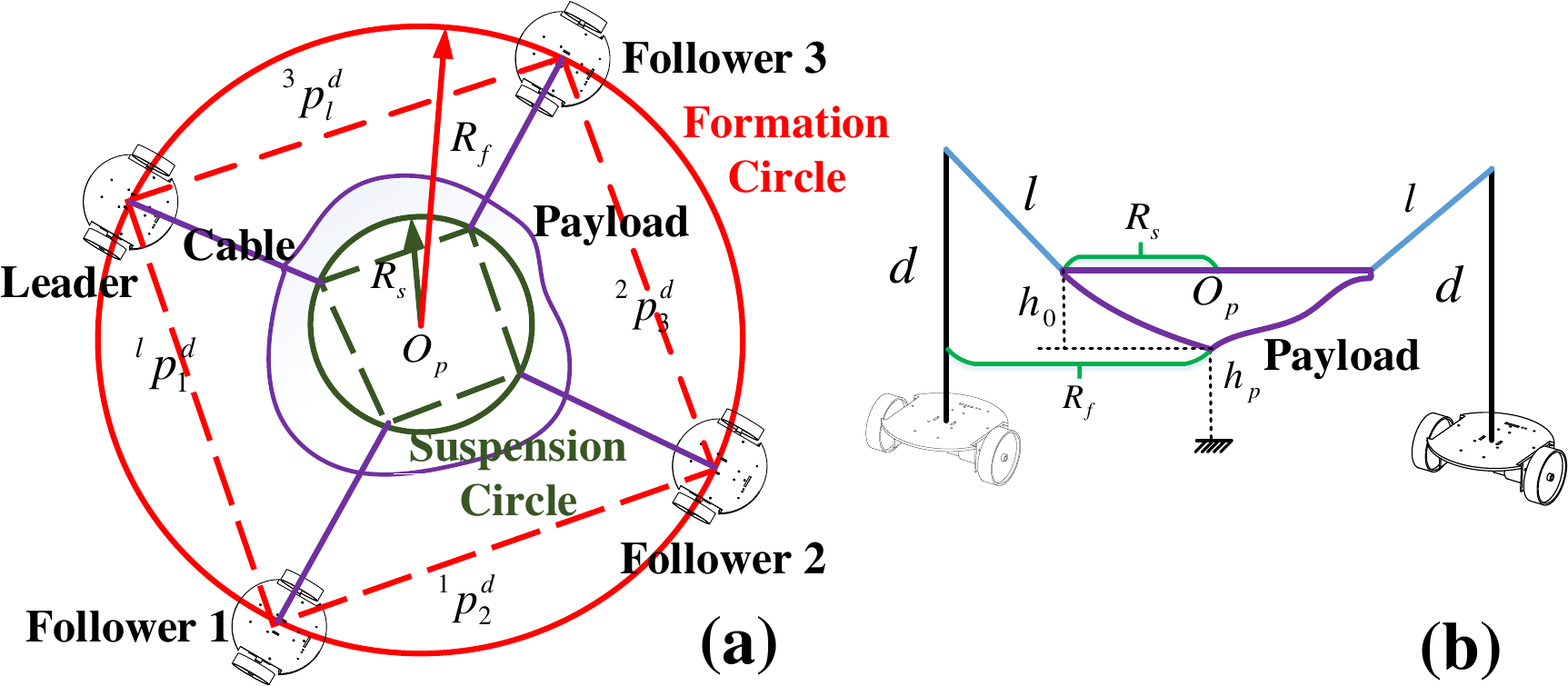}  
\caption{Top view and front view of the transportation system}
\label{TransportDetails}
\vspace{-0.2cm}
\end{figure} 

To make the force on robots uniform, we set the desired relative position between robots on a circle, called the formation circle, as in Fig. \ref{TransportDetails}. Besides, the polygon formed by suspension points on the payload is similar to the polygon formed by robots, whose circumscribed circle is named as the suspension circle. As for parameters, the radius of the suspension circle and the formation circle are denoted as $R_s$ and $R_f$ respectively, while the height of payload is $h_0$. 


\subsection{Constraints of Transportation System}
There are many constraints to be preserved during transportation. In the following, we will present some representative constraints and show how to scale up new constraints. 
\subsubsection{Payload Protection Constraints}
The payload may be stretched or collide with the ground. Assume the distance from the lowest point of payload to the ground as $h_p$ and the minimum safety height of payload as $h_{\min}$. To avoid collisions with the ground, there exists
\begin{equation} \label{eq200}
h_p=d-h_0-\sqrt{l^2-(R_f-R_s)^2}\ge h_{\min} 
\end{equation}
if we ignore the oscillation of the payload.

To prevent the payload from being stretched, $R_f$ should satisfy $R_f \le l+R_s$. Combing Equation (\ref{eq200}), we have 
\begin{align} \label{eq2}
\begin{array}{l}
R_{\min} \le R_f \le l+R_s\\
R_{\min}=\left\{\begin{array}{l}
R_s\,(k_0> l, \text{where}\; k_0 = d-h_0-h_{\min})\\
R_s+\sqrt{l^2-k_0^2}\;(k_0\le l)
\end{array}\right.
\end{array}
\end{align} 

Thus, the robot formation can be scaled between circles of radius $R_{\min}$ to $l+R_s$, where the maximum scale ratio is defined as $r_{+}=\frac{l+R_s}{R_f}$ and the minimum scale ratio is $r_{-}=\frac{R_{\min}}{R_f}$. To restrict the formation within this annulus district and avoid collisions, the relative distance $\Vert {}^i\boldsymbol{p}_{j} \Vert$ should satisfy the constraint in Equation (\ref{eq2a}), where $\Vert {}^i\boldsymbol{p}_j^d \Vert$ is the desired relative distance in the Cartesian coordinate system. In the following, $\Vert {}^i\boldsymbol{p}_j^d \Vert$ will be replaced by an approximation $\Vert {}^i\boldsymbol{c}_j^d \Vert$, which is the desired relative distance in the curvilinear coordinate system and defined later.
\subsubsection{Individual Collision Avoidance Constraints}

Each robot $i$ should keep a safety distance $d_o$ from nearby obstacles for collision avoidance as in Equation (\ref{eq2b}),
where $\rho_{io}$ is the distance between robot $i$ and its nearest obstacle, and $\mathcal{O}_i(t)$ is the obstacle measurement of robot $i$ at time $t$. Similarly, robots $i$ and $j$ should maintain at least $\rho_c$ away from each other to avoid inter-robot collisions as in Equation (\ref{eq2c}). 

\subsubsection{System Collision Avoidance Constraints}
In addition to robots, the cables and payload in the system should also be prevented from obstacles. As Fig. \ref{Transport} shows, they are restricted in the polygon generated by adjacent robots. If each edge $L_{ij}$ between robot $i$ and $j$ in the polygon keeps a safety distance $d_l$ away from obstacles, the system would not be damaged, represented as in Equation (\ref{eq2d}). For robot $i$, it only needs to consider collision avoidance between its two adjacent edges and obstacles and the two related robots form the set $\mathcal{Q}_i$. For example, the set $\mathcal{Q}_3$ for Follower $3$ in Fig. \ref{TransportDetails} is $\mathcal{Q}_3=\{l,2\}$. 

Besides, the function ${dist}({}^i\boldsymbol{p}_o,L_{ij}) = \frac{{}^i\boldsymbol{p}_o^TG{}^i\boldsymbol{p}_j}{\Vert G{}^i\boldsymbol{p}_j \Vert}$ is the distance from the obstacle $\boldsymbol{p}_o$ to $L_{ij}$ when ${}^i\boldsymbol{p}_o \in D_{ij}$, where ${}^i\boldsymbol{p}_o$ is the coordinate of $\boldsymbol{p}_o$ in the frame of robot $i$, $D_{ij}=\left\{q \mid\left(q-\boldsymbol{p}_{i}\right)^{T}{}^{i} \boldsymbol{p}_{j}>0,\left(q-\boldsymbol{p}_{j}\right)^{T} {}^{i} \boldsymbol{p}_{j}<0\right\}$ and the matrix $G$ is 
$\left[ \begin{array}{ccc}
0 & -1\\
1& 0
\end{array}
\right ]$.
Furthermore, $\rho_{il}$ is the smaller one of two available distances of $i$ and $d_l$ is the safety distance.   
\subsubsection{Velocity Constraints}
Due to mechanical and electrical constraints, the velocities of robot $i$ are usually bounded as in Equation (\ref{eq2e}), where $\bar v$ and $\bar \omega$ are velocity bounds.
\subsubsection{Sensor Range Constraints}
In general, cameras have a measuring range $[0,c_{\max}]$. In order for cameras on robots to observe each other's positions successfully, the distances between robots should be within $c_{\max}$ as in Equation (\ref{eq2f}).
\subsubsection{Other Scalable Constraints}
In practice, there may be other constraints on robot $i$. These constraints are usually related to velocity and position of robots and can be expressed uniformly as in Equation (\ref{eq2g}), where $\delta_i^-$ and $\delta_i^+$ are the lower and upper bounds, $f_i$ and $g_i$ are constraint functions. If the constraint is unbounded on one side, then $\delta_i^-$ is set to $-\infty$ or $\delta_i^+$ to $+\infty$. When the constraint is an equation constraint, we set $\delta_i^-$ and $\delta_i^+$ very close to each other for approximation.  

Based on above discussion, all constraints for the system can be summarized as follows. 
\begin{subequations}
\begin{align}
&r_{-}\Vert {}^i\boldsymbol{p}_{j}^d \Vert \le \Vert {}^i\boldsymbol{p}_{j} \Vert \le r_{+}\Vert {}^i\boldsymbol{p}_{j}^d \Vert \;(i\in \mathcal{F},j\in \mathcal{N}_i)\label{eq2a}\\
&\rho_{io} = \min _{\boldsymbol{p}_o \in \mathcal{O}_i(t)}\| \boldsymbol{p}_o- \boldsymbol{p}_{i}\| \ge d_o\;(i\in \mathcal{R}) \label{eq2b}\\
&\Vert {}^i\boldsymbol{p}_{j} \Vert \ge \rho_c\;(i\in \mathcal{F},j\in \mathcal{N}_i) \label{eq2c}\\
&\rho_{il} = \min_{\boldsymbol{p}_o \in \mathcal{O}_i(t);j\in \mathcal{Q}_i} \text{dist}({}^i\boldsymbol{p}_o,L_{ij}) \ge d_l \label{eq2d}\;(i\in \mathcal{F})\\
&\vert v_i \vert \le \bar v,\;\vert \omega_i \vert \le \bar \omega \;(i\in \mathcal{R})\label{eq2e}\\
&\Vert {}^i\boldsymbol{p}_{j} \Vert \le c_{\max}\;(i\in \mathcal{F},j\in \mathcal{N}_i) \label{eq2f}\\
&\delta_i^{-} \le f_i(v_i,\omega_i)+g_i(\boldsymbol{p}_i,\boldsymbol{p}_j)\le \delta_i^{+}\;(i\in \mathcal{F},j\in \mathcal{N}_i) \label{eq2g}
\end{align}
\end{subequations}
where only constraints in Equation (\ref{eq2b}), (\ref{eq2e}) are relevant for the leader navigating to $\boldsymbol{p}_g$ and other constraints related to the leader can be preserved by corresponding followers.    
\begin{Remark}
Here, we omit the motion and force analysis of the payload for the following reasons. First, accurately measuring the forces on the payload is difficult without using expensive force sensors, as some ropes may not always be under tension. Second, the current conditions account for worst-case positional scenarios, ensuring the payload’s safety, making further motion analysis less necessary.
\end{Remark}

\begin{figure*}[htbp]
\centering
\includegraphics[height=1.4in]{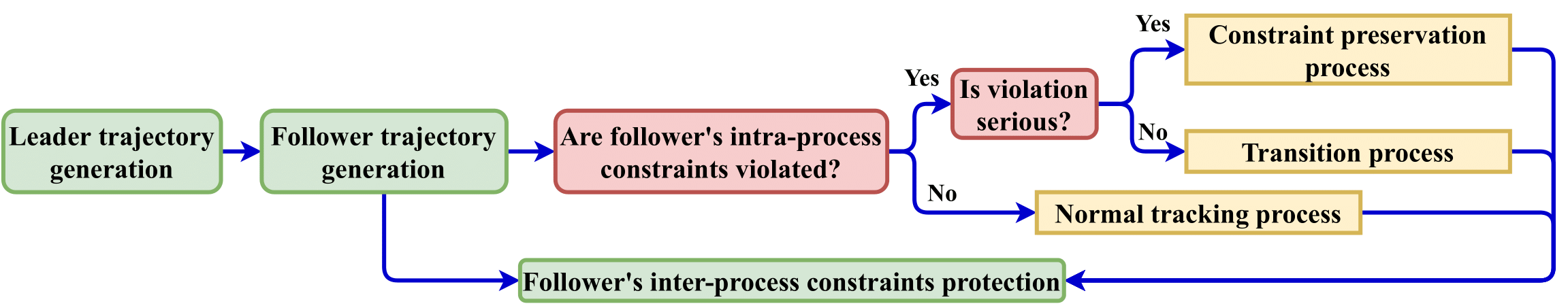}  
\caption{Flowchart of transportation framework, where yellow boxes represent robot process and red boxes are judgment conditions}
\label{flowchart} 
\vspace{-0.7cm}
\end{figure*}
\subsection{Problem Description}
Given the assumption that obstacles are convex and there is sufficient space for the transportation system with above constraints to pass through, we obtain the following problem. 
\begin{Problem} \label{prob_1}
Consider a cooperative transportation system comprised of a cable-suspended payload and several nonholonomic mobile robots under above constraints. Given a final goal position $\boldsymbol{p}_g=[x_g,y_g]^T$, we should propose distributed control strategies to achieve  
\begin{enumerate}
\item The reference trajectories to $\boldsymbol{p}_g$ can be quickly generated for robots without relying on global maps and time-consuming trajectory optimization algorithms.
\item Robots can overcome scalable constraints while tracking reference trajectories, and the time consumed is only linearly related to the number of robots and constraints.
\end{enumerate}          
\end{Problem} 
\subsection{Main Structure of Proposed Scheme}
The main structure of our proposed approach to Problem \ref{prob_1} is shown in Fig. \ref{flowchart}. As shown, the leader robot first quickly generates a trajectory towards $\boldsymbol{p}_g$, and then followers generate their own reference trajectories based on the leader's. During transportation, it is mainly the follower who is responsible for overcoming system constraints. The states of followers are divided into constraint preservation process, transition process and normal tracking process. The majority of constraints can be preserved in constraint preservation process, named as intra-process constraints. However, there are a few constraints that exist at all times and need to be protected in all processes such as velocity constraints, named as inter-process constraints.
These algorithms and processes will be described in details in the following sections. 
\section{Cooperative Trajectory Generation}
In this section, the methods of generating reference trajectory for the leader and followers are presented. Unlike other methods, ours takes only a constant amount of time $O(1)$ to complete at each sampling instant without global maps.
\begin{figure}[htbp]
\vspace{-0.2cm}
\centering
\includegraphics[width=2.4in]{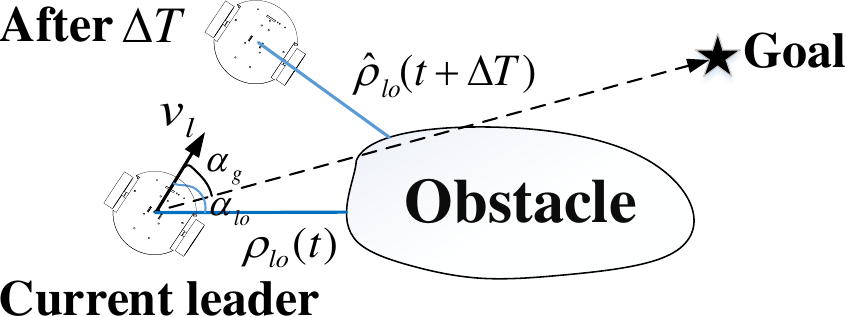}  
\caption{The diagram of leader trajectory generation}
\label{leader_graph}
\vspace{-0.8cm}
\end{figure} 
\subsection{Leader Trajectory Generation}
The leader is responsible for guiding the system to $\boldsymbol{p}_g$ and selecting a path which is spacious enough for robots to pass through. To facilitate it, followers are preferably distributed evenly on either side of the leader. With obstacle data $\mathcal{O}_l$ from the laser scanner, the trajectory generation algorithm of the leader is proposed in Algorithm \ref{algorithm_1}. As in Fig. \ref{leader_graph}, the nearest obstacle point $\boldsymbol{p}^{*}_o=[{x}^{*}_o, {y}^{*}_o]^T$ is first calculated in the leader frame with $\mathcal{O}_l$. Then, we define the distance from the leader to $\boldsymbol{p}^{*}_o$ as $\rho_{lo}=\sqrt{({x}^{*}_o)^2+({y}^{*}_o)^2}$ and the relative angle as $\alpha_{lo}=\text{atan2}(y_o^*,x_o^*)$, while $\alpha_g$ is the relative angle to the goal $\boldsymbol{p}_g$, $atan2$ is the function to calculate the arctangent value, $F$ is a flag variable to record the sign of $\dot{\rho}_{lo}$. The sign function ${sgn}(x)$ is defined as $\frac{x}{\vert x \vert}$ when $x \neq 0$, and $0$ otherwise.  

When $\rho_{lo} > d_t$, the leader is far from obstacles. It moves forward and adjusts the heading angle towards $\boldsymbol{p}_g$. The threshold $d_t$ should be large enough to cover the passage of the whole system, e.g. $d_t>R_f$. If $\rho_{lo} \le d_t$, the leader makes decisions according to $F$. When $F>0$, the leader is heading away from obstacles and it should continue moving towards $\boldsymbol{p}_g$. If $F\le 0$, a turn is performed to steer the leader away from obstacles. Specifically, it turns in the positive direction when the nearest obstacle is on the left, and in the negative direction otherwise. The linear velocity of leader is designed as ${v}_l=\varphi(\rho_{lo})\frac{}{}\tilde{v}_l$, which decreases as $\rho_{lo}$, i.e.
\begin{align} \label{eq51}
\varphi(\rho_{lo})=\left\{\begin{array}{l}
1\;(\rho_{lo}\ge d_t);\;0\;(\rho_{lo}<d_{o})\\
\frac{\rho_{lo}-d_{o}}{d_t-d_{o}}\;(d_{o} \le \rho_{lo}< d_t)
\end{array}\right.
\end{align}  
where $\tilde{v}_l$ is a constant. When approaching obstacles, it is beneficial for the leader to slow down to avoid them.

\vspace{-0.3cm}
\begin{algorithm} \label{algorithm_1}
\caption{Leader Trajectory Generation}
\KwOut{$v_l$, $\omega_l$}
Obtain the nearest obstacle $\boldsymbol{p}^{*}_o=\text{argmin}_{\boldsymbol{p}\in \mathcal{O}_l}(\rho_{lo})$ in the local frame of the leader\; 
Calculate obstacle distance $\rho_{lo}$ and relative angle $\alpha_{lo}$\;
$\alpha_g = \operatorname{atan2}(y_g-y_l,x_g-x_l)-\theta_l$\;
$F = \text{sgn}(\hat \rho_{lo}(t+\Delta T)-\rho_{lo}(t))$, $v_l = \varphi(\rho_{lo})\frac{}{}\tilde{v}_l$\;
\uIf{$\rho_{lo} > d_t$}
{
    $\omega_l = \tilde{\omega}_l \text{sgn}(\alpha_g)$\; 

}\uElseIf{$\rho_{lo} \le d_t$ and $F>0$}
{
    $\omega_l = \tilde{\omega}_l \text{sgn}(\alpha_g)$\;
}\uElseIf{$\rho_{lo} \le d_t$, $F\le 0$ and $\alpha_{lo}<0$}
{
    $\omega_l = \tilde{\omega}_l$\;
}\ElseIf{$\rho_{lo} \le d_t$, $F\le 0$ and $\alpha_{lo}\ge 0$}
{
    $\omega_l = -\tilde{\omega}_l$\;
}
\end{algorithm}
\vspace{-0.3cm}

To obtain $F$ at time $t$, the difference between the estimation $\hat \rho_{lo}(t+\Delta T)$ and $\rho_{lo}(t)$ is applied, where $\Delta T$ is the forward prediction time. Besides, $\hat \rho_{lo}(t+\Delta T)$ is
\begin{equation} \label{eq6}
\begin{array}{l}
\hat \rho_{lo}(t+\Delta T) = \min _{\boldsymbol{p} \in \mathcal{O}_l(t)}\| \boldsymbol{p}- \hat{\boldsymbol{p}}_l(t+\Delta T)\|\\
\hat{\boldsymbol{p}}_l(t+\Delta T) = \boldsymbol{p}_l(t) + \Delta T[{v}_l \cos\theta_l(t), v_l \sin\theta_l(t)]^T
\end{array}
\end{equation}
Also, leader velocities are required to satisfy constraints in Equation (\ref{eq2e}), which means $\tilde{v} \le \bar v$ and $\tilde{\omega} \le \bar \omega$. To ensure obstacle avoidance in Equation (\ref{eq2b}), parameters should satisfy $\frac{\tilde{v}}{\tilde{\omega}}<d_o<d_t-\frac{2\tilde{v}}{\tilde{\omega}}<d_r-\frac{2\tilde{v}}{\tilde{\omega}}$,
where $d_r$ is the sensing range of the laser scanner. With above parameters, the leader would advance to $\boldsymbol{p}_g$ automatically and keep the distance $d_t-\frac{2\tilde{v}}{\tilde{\omega}}$ away from obstacles. To leave enough space for followers, we can increase $d_t$, $\tilde{\omega}$ and decrease $\tilde{v}$. The proof is similar to Theorem 1 in \cite{A.S.2013}, provided for interested readers to refer to.
\subsection{Follower Trajectory Generation}          \label{trajectory_generation}
Here, the reference trajectories are generated for followers. The rigid formation based on Cartesian coordinates suffers from many limitations and even a square shape with four nonholonomic mobile robots can not be perfectly achieved. To solve it, we use curvilinear coordinates instead. As in Fig. \ref{Transport}, the relative position between robots $i$ and $j$ is described by ${}^is_j$ and ${}^iq_j$, where ${}^is_j$ is the displacement along the path of robot $i$, and ${}^iq_j$ is the offset in the sliding direction. Notice that ${}^i\boldsymbol{c}_j = [{}^is_j,{}^iq_j]^T$ is equal to ${}^i\boldsymbol{p}_j$ when robot $i$ moves forward without rotation. Define the desired value of ${}^i\boldsymbol{c}_j$ as ${}^i\boldsymbol{c}_j^d = [{}^is_j^d,{}^iq_j^d]^T$. We generate the reference trajectory for the follower $i$ based on ${}^l\boldsymbol{c}_i^d$ with the leader $l$ in Algorithm \ref{algorithm_2}.
\vspace{-0.3cm}
\begin{algorithm} \label{algorithm_2}
\caption{Trajectory Generation for Follower $i$}
\If{$\vert \delta_{i} \vert > \delta_c$}
{
    Rotate its camera with $\omega_i^c = k_c(\delta_{i}-\delta_c \text{sgn}(\delta_{i}-\delta_{c}))$ to measure relative pose ${}^i\bm{m}_{l}$ to the leader\; 
}
Receive part of leader trajectory $\mathcal{T}_l(t)$ from the leader\;
Calculate its desired position $O_i^d(t)$ in the leader frame, which is $({}^is_j^d,{}^iq_j^d)$ away from the leader on $\mathcal{T}_l(t)$ in curvilinear coordinate systems\;  
Transform $O_i^d$ from the leader frame to the local frame of follower $i$ based on ${}^i\bm{m}_{l}$\;
Obtain its reference velocity $({}^iv_r(t),{}^i\omega_r(t))$ according to Equation (\ref{eq8})\; 
%
\end{algorithm}
\vspace{-0.3cm}

To obtain continuous relative pose measurements ${}^i\bm{m}_{l}=[{}^i\bm{p}_{l}^T,{}^i\theta_{l}]^T$ to the leader, the limited-FOV camera on follower $i$ actively tracks the leader via a rotating motor below, where the leader is detected by AprilTag in \cite{J.W.2016}. When the absolute value of relative angle $\delta_i$ between the camera and the leader exceeds the threshold $\delta_c$, the camera rotates at $\omega_i^c$ to track the leader and maintain it in FOV.
Then, the leader transmits part of its trajectory $\mathcal{T}_l(t)$ to all followers, where $\mathcal{T}_l(t)$ records historical poses and velocities of the leader. Specifically, $\mathcal{T}_l(t)$ is defined as $\mathcal{T}_l(t)=(\bm{g}_l(t_{-k}),\bm{g}_l(t_{-k+1}),...,\bm{g}_l(t_{-1}),\bm{g}_l(t))$, where $\bm{g}_l(t)=[\bm{p}_l(t), \theta_l(t), v_l(t),\omega_l(t)]^T$ is recorded in the leader frame, $t_p$ represents $p^{\text{th}}$ sampling instant and $k$ is the buffer length. To reduce the amount of communication, we can decrease the buffer length $k$ as much as possible. Besides, data are transmitted only when leader velocities change, because the leader poses can be inferred from the constant leader velocities when they are unchanged.  

Next, as Fig. \ref{Transport} shows, the follower $i$ calculates the position of $O_L^i$ in its own frame with $\mathcal{T}_l(t)$, which is ${}^ls_i^d$ away from the current position of the leader $O_L$ along the leader's path. The time instant $t_a$ when the leader arrives at $O_L^i$ is estimated by $t_a = \text{argmin}_t\{\text{dist}(O_L^i, \bm{p}_l(t))\}$. Next, construct the coordinate system at $O_L^i$ with the forward tangent line of the leader's path as the $x$ axis. The desired position $O_i^d$ for the follower $i$ is ${}^lq_i^d$ away from $O_L^i$ on the $y$ axis and the desired heading angle $\theta_i^d$ is parallel to the $x$ axis. After obtaining $O_i^d$, the follower $i$ transforms $O_i^d$ to its own local frame based on ${}^i\bm{m}_{l}$.


The reference velocities ${}^iv_r(t)$ and ${}^i\omega_r(t)$ of the follower $i$ are generated with leader velocities  $(v_l(t_a),\omega_l(t_a))$ at $t_a$ as 
\begin{equation} \label{eq8}
{}^iv_r(t)=v_l(t_a)-\omega_l(t_a){}^lq_i^d,\;{}^i\omega_r(t)=\omega_l(t_a)
\end{equation}
where ${}^iv_r(t)$ is the sum of the leader velocity $v_l(t_a)$ and the additional velocity ${}^l\omega_i(t_a){}^lq_i^d$ caused by leader rotation, and ${}^i\omega_r(t)$ remains the same as the leader to maintain consistency.
\subsection{Complexity Analysis}
When the leader executes Algorithm \ref{algorithm_1}, the most time-consuming operation is the calculation of $\rho_{lo}(t)$ and $\hat \rho_{lo}(t+\Delta T)$, which requires traversing all points in $\mathcal{O}_l(t)$ and takes the time related to the number of points $y$ in $\mathcal{O}_l(t)$. The remaining parts are elementary operations, so the leader performs about $2y$ steps to complete Algorithm \ref{algorithm_1}. In fact, $y$ is a constant, so the leader only consumes a constant time $O(1)$. For each follower $i$, the most time-consuming operation is the calculation of $O_i^d$ and $t_a$, which requires traversing the received leader trajectory, and the steps performed are related to the buffer length $k$. Since $k$ is an adjustable constant, the time taken by the follower is also a constant $O(1)$. 

In contrast, integer programming based multi-robot trajectory generation methods are NP-hard with $O(2^n)$ complexity, while quadratic programming based optimization methods require a polynomial complexity $O(n^m)$, where $n$ is the number of constraints or robots. In addition, they generate trajectories offline based on global maps, which makes them difficult to extend to new environments and not real-time. Fortunately, these shortcomings can be overcome by our approach.

\section{Time-efficient Cooperative Transportation under Scalable Constraints}
As in Fig. \ref{flowchart}, the follower is required to track the generated trajectory and maintain constraints based on three designed processes. In the normal tracking process, the follower has no tendency to break any intra-process constraints. It tracks the trajectory normally and preserves inter-process constraints like velocity constraints. When any intra-process constraints are about to break, it comes into the constraint preservation process and actions are taken to avoid them without considering the tracking mission. Otherwise, the follower is in the transition process, where the tracking mission and constraint maintenance are considered concurrently. The three processes and the associated algorithms are discussed as follows.
\subsection{Normal Tracking Process}
Based on \cite{X.Y.2015}, the control strategies for the follower $i$ in the normal tracking process are designed as
\begin{equation}
\nonumber
\begin{aligned}
v_i^n &={}^iv_{r}+\frac{c_{1} {}^ix_{e}}{\sqrt{1+{}^ix_{e}^{2}+{}^iy_{e}^{2}}} \\
\omega_i^n &={}^i\omega_{r}+\frac{c_{2} {}^iv_{r}({}^iy_{e} \cos \frac{{}^i\theta_{e}}{2}-{}^ix_{e} \sin \frac{{}^i\theta_{e}}{2})}{\sqrt{1+{}^ix_{e}^{2}+{}^iy_{e}^{2}}}+c_{3} \sin \frac{{}^i\theta_{e}}{2}
\end{aligned}
\end{equation}
where $c_1$, $c_2$ and $c_3$ are positive parameters. The tracking errors are ${}^ix_{e}=\Vert {}^i\bm{p}_e \Vert \cos ({}^i\alpha_e-\theta_i)$, ${}^iy_{e}=\Vert {}^i\bm{p}_e \Vert \sin ({}^i\alpha_e-\theta_i)$ and ${}^i\theta_{e}=\theta_i^d-\theta_i$ in the local frame of follower $i$, while $\Vert {}^i\bm{p}_e \Vert=\sqrt{(x_i^d-x_i)^2+(y_i^d-y_i)^2}$, ${}^i\alpha_e =atan2(y_i^d-y_i,x_i^d-x_i)$, $[x_i^d,y_i^d]^T$ is the position of $O_i^d$ and reference velocities ${}^iv_{r}$, ${}^i\omega_{r}$ are obtained from Equation (\ref{eq8}). 

Then, we discuss the velocity constraint in Equation (\ref{eq2e}). The items in $v_i^n$ and $\omega_i^n$ satisfy $\vert \frac{c_{1} {}^ix_{e}}{\sqrt{1+{}^ix_{e}^{2}+{}^iy_{e}^{2}}}\vert \le c_1$ and $\vert\frac{c_{2} ({}^iy_{e} \cos \frac{{}^i\theta_{e}}{2}-{}^ix_{e} \sin \frac{{}^i\theta_{e}}{2})}{\sqrt{1+{}^ix_{e}^{2}+{}^iy_{e}^{2}}}\vert \le \vert \frac{c_{2} \sqrt{{}^ix_{e}^{2}+{}^iy_{e}^{2}}}{\sqrt{1+{}^ix_{e}^{2}+{}^iy_{e}^{2}}}\vert\le c_2$. 
To preserve the constraint, we have $\vert v_i \vert \le\vert {}^iv_{r}\vert +c_1\le \tilde v+\tilde \omega \vert {}^lq_j^d \vert + c_1\le \bar v$ and $\vert \omega_i\vert \le \tilde \omega+c_2 \tilde v + c_3 \le \bar\omega$.
\subsection{Constraint Preservation Process}
Next, the constraint preservation issue is considered. To achieve it for the follower $i$, three different kinds of intra-process constraints are discussed as follows.
\subsubsection{Two-sided Constraints} 
These constraints have both maximum and minimum values, such as Equation (\ref{eq2a}) and constraints merged by Equation (\ref{eq2c}), (\ref{eq2f}). We can rewrite them uniformly as $\delta_{a}^{i-}\le g_a^i(\bm{p}_i,\bm{p}_j) \le \delta_{a}^{i+}\,(j\in \mathcal{N}_i)$, where $\delta_{a}^{i-}$ and $\delta_{a}^{i+}$ are the corresponding lower and upper bounds. To maintain them, the function $\mathcal{Q}_{a}^i(g_a^i)$ is designed as  
\begin{equation}
\nonumber
\mathcal{Q}_{a}^i(g_a^i)=\left\{\begin{array}{l}
\frac{(g_a^i-\delta_{a}^{i-}-\gamma_a^{i})^2}{-\delta_{a}^{i-}+g_a^i+(\gamma_a^{i})^2}\;(\delta_{a}^{i-} \le g_a^i \le \delta_{a}^{i-} +\gamma_a^{i})\\
0\;(\delta_{a}^{i-} + \gamma_a^{i}< g_a^i \le \delta_{a}^{i+} - \gamma_a^{i})\\
\frac{(g_a^i-\delta_{a}^{i+}+\gamma_a^{i})^2}{\delta_{a}^{i+}-g_a^i+(\gamma_a^{i})^2}\;(\delta_{a}^{i+} - \gamma_a^{i} < g_a^i \le \delta_{a}^{i+})
\end{array}\right.
\end{equation}
where $\gamma_a^{i}<\frac{\delta_{a}^{i+}-\delta_{a}^{i-}}{2}$ is a positive parameter to adjust the available range of $g_a^i$. We define the function for the $k^{th}$ two-sided constraint of the follower $i$ as $\mathcal{Q}_{a}^{ik}$, the corresponding variable as $g_a^{ik}$, the parameter as $\gamma_a^{ik}\,(k\in \{1,2,...,q_a\})$ and the number of two-sided constraints as $q_a$.  
\subsubsection{One-sided Constraints}
These constraints are bounded only on one side, such as constraints in Equation (\ref{eq2b}), (\ref{eq2d}). They are summarized as two types, including $g_b^i(\bm{p}_i,\bm{p}_j) \ge \delta_{b}\,(j\in \mathcal{N}_i)$ and $g_c^i(\bm{p}_i,\bm{p}_j) \le \delta_{c}\,(j\in \mathcal{N}_i)$, where $\delta_{b}$ and $\delta_{c}$ are the lower and upper bounds. Then, the functions corresponding to these two types are respectively designed as
\begin{equation}
\nonumber
\mathcal{Q}_b^i(g_b^i)=\left\{\begin{array}{l}
\frac{(g_b^i-\delta_{b}^{i}-\gamma_b^{i})^2}{-\delta_{b}^{i}+g_b^i+(\gamma_b^{i})^2}\;(\delta_{b}^{i} \le g_b^i < \delta_{b}^{i} + \gamma_b^{i})\\
0\;(g_b^i  \ge \delta_{b}^{i} + \gamma_b^{i})
\end{array}\right.
\end{equation}
\begin{equation}
\nonumber
\mathcal{Q}_c^i(g_c^i)=\left\{\begin{array}{l}
\frac{(g_c^i-\delta_{c}^{i}+\gamma_c^{i})^2}{\delta_{c}^{i}-g_c^i+(\gamma_c^{i})^2}\;(\delta_{c}^{i} - \gamma_c^{i} < g_c^i \le \delta_{c}^{i})\\
0\;(g_b^i  \le \delta_{c}^{i} - \gamma_c^{i})
\end{array}\right.
\end{equation}
where $\gamma_b^{i}$ and $\gamma_c^{i}$ are adjustable parameters. Similarly, we define the function for the $k^{th}$ left-bounded constraint as $\mathcal{Q}_b^{ik}$, while $\mathcal{Q}_c^{ik}$ is for the $k^{th}$ right-bounded constraint. The corresponding parameters are $\gamma_b^{i}$ and $\gamma_c^{i}$, while the corresponding variables are  $g_b^{ik}$ and $g_c^{ik}$. Besides, the number of these two kinds of constraints are $q_b$ and $q_c$, relatively. 
\subsubsection{Mixed Constraints}
There are some constraints that may be mixed with velocities like Equation (\ref{eq2g}), which are related to inter-process constraints. 
We assume that the upper bound of $f_i$ can be inferred from Equation (\ref{eq2e}). For example, $f_i=v_i^2+v_i\omega_i$ satisfies our assumptions, because it is deduced that $\vert f_i \vert \le \bar v^2+\bar v\bar \omega$. Instead, $f_i=\frac{1}{v_i}$ is not suitable, because the upper bound of $f_i$ can not be obtained from $\vert v_i \vert \le \bar v$.
Suppose the upper bound of $f_i$ calculated from Equation (\ref{eq2e}) as $\vert f_i \vert \le \bar f$. We can decouple two functions $f_i$, $g_i$ and obtain 
\begin{equation} \label{eq4a}
\bar f+\delta_i^- \le g_i(\boldsymbol{p}_i,\boldsymbol{p}_j) \le \delta_i^+-\bar f    
\end{equation}
as a new intra-process constraint to maintain. To achieve it, $\bar f $ should satisfy $\bar f \le \frac{1}{2}(\delta_i^+-\delta_i^-)$, which can be decreased by lowering $\bar v$ and $\bar \omega$ when Equation (\ref{eq4a}) does not hold.

In general, the above three functions approach the maximum value $1$ when constraints tend to be avoided, while they remain the minimum $0$ if constraints are protected. Define $\beta_i$ as the maximum value of all related functions for follower $i$, we have 
\begin{equation}
\beta_i = \max\{\mathcal{Q}_{a}^{ik_a}\,, \mathcal{Q}_b^{ik_b}, \mathcal{Q}_c^{ik_c}\} \in [0,1]
\end{equation}
where $k_a$, $k_b$ and $k_c$ are the indexes of functions. 

The process that the follower belongs to is recognized by $\beta_i$. When $\beta_i=0$, the follower $i$ is in the normal tracking process. If $\beta_i$ is larger than the threshold $\beta_t \in (0,1)$, the follower $i$ enters the constraint preservation process. Otherwise, it moves into the transition process. Then, the multi-objective control strategies for followers under constraints are summarized in Algorithm \ref{algorithm_3}. Here, the saturation function ${sat}(x,y)$ is equal to $1$ when $|x|\le y$ and ${sgn}(x) \frac{y}{x}$ otherwise. Obviously, the function ${sat}(x,y)$ is always greater than or equal to zero. The vector $ \tau_i=\nabla_{\bm{p}_i} \Psi_i=[ \nabla_x \Psi_i, \nabla_y \Psi_i]^T$ is the gradient of $\Psi_i$ with respect to $\bm{p}_i$. Next, the following theorem is obtained. 
\vspace{-0.3cm}
\begin{algorithm} \label{algorithm_3}
\caption{Follower $i$'s Strategy with Constraints}
\KwOut{$v_i$, $\omega_i$}
$\mathcal{P}_{a}^{ik_a}=1-\mathcal{Q}_{a}^{ik_a},\mathcal{P}_b^{ik_b}=1-\mathcal{Q}_b^{ik_b},\mathcal{P}_c^{ik_c}=1-\mathcal{Q}_c^{ik_c}$\;
$\Psi_i = 1-(\prod_{k_a=1}^{q_a}\mathcal{ P}_{a}^{ik_a}) (\prod_{k_b=1}^{q_b}\mathcal{ P}_{b}^{ik_b})(\prod_{k_c=1}^{q_c}\mathcal{ P}_{c}^{ik_c})$\;
$\tau^i = [\tau_x^i,\tau_y^i]^T=[\nabla_x \Psi_i, \nabla_y \Psi_i]^T$\;
$\phi_i^d = \text{atan2}(\tau_y^i,\tau_x^i)$\;
$\hat v_i^o = -k_v^o\cos(\theta_i-\phi_i^d)\sqrt{(\tau_x^i)^2+(\tau_y^i)^2}$\;
$\hat \omega_i^o=-k_\omega^o(\theta_i-\phi_i^d)$\;
$v_i^o = \text{sat}(\hat{v}_i^o, \bar v)\hat{v}_i^o$, $\omega_i^o = \text{sat}(\hat{\omega}_i^o, \bar \omega)\hat{\omega}_i^o$\;
\uIf{$\beta_{i} = 0$}
{
    $v_i = v_i^n$, $\omega_i = \omega_i^n$\;  

}\uElseIf{$0<\beta_{i}\le \beta_t$}
{
    $v_i = \frac{\beta_t-\beta_i}{\beta_t}v_i^n + \frac{\beta_i}{\beta_t}v_i^o$, $\omega_i = \frac{\beta_t-\beta_i}{\beta_t}\omega_i^n + \frac{\beta_i}{\beta_t}\omega_i^o$\;
}\ElseIf{$\beta_t < \beta_{i} \le 1$}
{
    $v_i = v_i^o$, $\omega_i = \omega_i^o$\;
}
\end{algorithm}
\vspace{-0.3cm}
\begin{Theorem}
For each follower $i$ in the system, it can preserve all intra-process constraints, i.e. $\beta_i \le 1$ and inter-process constraints on velocities in Equation (\ref{eq2e}), if strategies in Algorithm \ref{algorithm_3} are applied and $\beta_i\le \beta_t$ is satisfied initially.    
\end{Theorem}
\begin{Proof}
In the normal tracking process and the transition process, there exists $\beta_i \le \beta_t < 1$, which means that $\mathcal{Q}_{a}^{ik_a}<1$, $\mathcal{Q}_{b}^{ik_b}<1$ and $\mathcal{Q}_{c}^{ik_c}<1$. As designed, the constraints will not be avoided when the function $\mathcal{Q}^i$ is less than $1$. Thus, if we can show that these constraints are also kept during the constraint preservation process, the theorem would be proved.

First, the dangerous region $D_d^i$ is defined as the set of positions that cause the follower $i$ into the constraint preservation process. For the mobile robot system $\dot{\bm{p}}_i = \bm{u}_s^i$ with the single integrator model, there always exists $\Psi_i(\bm{p}_i(t))>0$ when $\bm{p}_i \in D_d^i$ and $\bm{u}_s^i$ is $-\tau^i$, based on the definition of $\Psi_i$. 

Consider a Lyapunov function candidate $V_i=\Psi_i$ for $\dot{\bm{p}}_i = \bm{u}_s^i$. The time derivative of $V_i$ is
\begin{equation} \label{eq13}
\dot V_i = (\nabla_{\bm{p}_i} \Psi_i)^T\dot{\bm{p}}_i = -(\nabla_{\bm{p}_i} \Psi_i)^T\nabla_{\bm{p}_i} \Psi_i \le 0.    
\end{equation}
Assume the instant when the follower $i$ comes into $D_d^i$ as $t_0$, there exists $V_i(t_0) = \Psi_i(t_0) \le 1-(1-\beta_t)^{q_a+q_b+q_c}$ when $\bm{p}_i \in D_d^i$, due to $\beta_i > \beta_t$. By combining Equation (\ref{eq13}), we have $\Psi_i(t) \le \Psi_i(t_0) \le 1-(1-\beta_t)^{q_a+q_b+q_c} < 1$ for arbitrary instant $t$ in the constraint preservation process. When any intra-process constraint is totally violated, $\Psi_i(t)$ is larger than $1$.   

Since we assume that there is enough space in the environment for the transportation system to pass through, there exists an attraction region $\Omega_i$ for the robot system $\dot{\bm{p}}_i = -\tau^i$, such that the robot in $\Omega_i$ can eventually drive $\Psi_i$ to the minimum value 0. Based on Theorem 4 in \cite{S.Z.2018}, the control law $v_i = v_i^o$ and $\omega_i = \omega_i^o$ causes the system with the nonholonomic model in Equation (\ref{eq1}) to converge into the same attraction region $\Omega_i$ as the system $\dot{\bm{p}}_i = -\tau^i$. Thus, $\Psi_i(t) < 1$ holds for the follower $i$ if Algorithm \ref{algorithm_3} is applied.

Besides, $v_i^o$ and $\omega_i^o$ in the constraint preservation process are saturated within the upper bounds $\bar v$ and $\bar \omega$, which means that $|v_i^o|\le \bar v$ and $|\omega_i^o|\le \bar \omega$. Thus, the inter-process constraints on velocities are also maintained when $\beta_t < \beta_i \le 1$.
\end{Proof}

Next, we discuss how to calculate related gradients. Specifically, $\nabla_{\boldsymbol p_i}\mathcal{Q}^i(\Vert {}^i\bm{p}_j \Vert)=\frac{d \mathcal{Q}^i(\Vert {}^i\bm{p}_j \Vert)}{d (\Vert {}^i\bm{p}_j \Vert)}\frac{{}^i\bm{p}_j}{\Vert {}^i\bm{p}_j \Vert}$, $\nabla_{\boldsymbol p_i}\mathcal{Q}^i(\rho_{io})=\frac{d \mathcal{Q}^i(\rho_{io})}{d \rho_{io}}\frac{{}^i\bm{p}_o}{\rho_{io}}$ and $\nabla_{\boldsymbol p_i}\mathcal{Q}^i(\rho_{il})=\frac{d \mathcal{Q}^i(\rho_{il})}{d \rho_{il}}\nabla_{\boldsymbol p_i}\rho_{il}$, where $\mathcal{Q}^i$ is an arbitrary $\mathcal{Q}$ function of $i$ and $\nabla_{\boldsymbol p_i}\rho_{il}$ is $\nabla_{\boldsymbol p_i }\rho_{il}=\frac{\nabla_{\boldsymbol p_i }({}^i\boldsymbol{p}_oG{}^i\boldsymbol{p}_j)\Vert G{}^i\boldsymbol{p}_j \Vert-{}^i\boldsymbol{p}_oG{}^i\boldsymbol{p}_j\nabla_{\boldsymbol p_i }(\Vert G{}^i\boldsymbol{p}_j \Vert)}{\Vert G{}^i\boldsymbol{p}_j \Vert^2}
=\frac{\Vert {}^i\boldsymbol{p}_j \Vert G{}^j\boldsymbol{p}_o+\rho_{il}{}^i\boldsymbol{p}_j}{\Vert {}^i\boldsymbol{p}_j \Vert^2}$.  
    
\subsection{Transition Process}
As for the transition process, the follower ought to balance trajectory tracking and constraint preservation concurrently. Here, we design a multi-objective control law based on $\beta_i$, which describes the risk of breaking intra-process constraints. When $\beta_i$ is close to $0$, it means that the follower operates safely and velocities for trajectory tracking mission should take up a larger proportion. Otherwise, if $\beta_i$ approaches to $\beta_t$, $v_i^o$ and $\omega_i^o$ would become main components. To make velocities continuous at $\beta_i = 0$ and $\beta_i = \beta_t$, the coefficients of $v_i^n$ and $v_i^o$ are designed as $\frac{\beta_t-\beta_i}{\beta_t}$ and $\frac{\beta_i}{\beta_t}$ relatively, and the same as $\omega_i^n$ and $\omega_i^o$. Furthermore, since $|\frac{\beta_t-\beta_i}{\beta_t}| + |\frac{\beta_i}{\beta_t}| = 1$, we can conclude that the inter-process constraints on robot velocities during the transition process are satisfied as well.  
\subsection{Complexity Analysis}
The most time-consuming operation in Algorithm \ref{algorithm_3} is the calculation of the gradient $\tau^i$ of $\Psi_i$. It is $\tau^i=-\sum_{k=1}^{q_a+q_b+q_c} \nabla_{\boldsymbol p_i}\mathcal{P}^{ik}\frac{1-\Psi_i}{\mathcal{P}^{ik}}$, where $\mathcal{P}^{ik}$ is an arbitrary $\mathcal{P}$ function for the $k^{th}$ constraint of follower $i$ and the total number of constraints is $n=q_a+q_b+q_c$. Normally, the number $n$ is larger than the number of neighboring robots $n_r$. If we can calculate in advance the expression for each gradient $\nabla_{\boldsymbol p_i}\mathcal{P}^{ik}$, the calculation of $\tau^i$ will become a combination of basic operations, which are repeated $n$ times. Thus, we can roughly consider the complexity as $O(n)$.


In contrast, other methods like hierarchical quadratic programming (HQP) suffer from higher complexity. As shown in \cite{D.K.2021}, if each robot has $n$ constraints, $n_r$ robots have roughly $\frac{nn_r}{2}$ constraints in total after removing duplicates. Based on \cite{E.A.2014}, HQP based on centralized optimization have a complexity of at least $O(m^2n_r)+O(n_r^2m)$, where $m$ is the number of constraints. Here, $m$ is $\frac{nn_r}{2}$ and the final complexity of HQP is about $O(n^5)$, worse than our methods with $O(n)$.       

\section{Simulation and Experiment} 
\subsection{Numerical Simulation} 
\begin{figure} [htbp]
\vspace{-0.5cm}
\begin{minipage}{0.5\linewidth}
\centering
\includegraphics[height=1.42in]{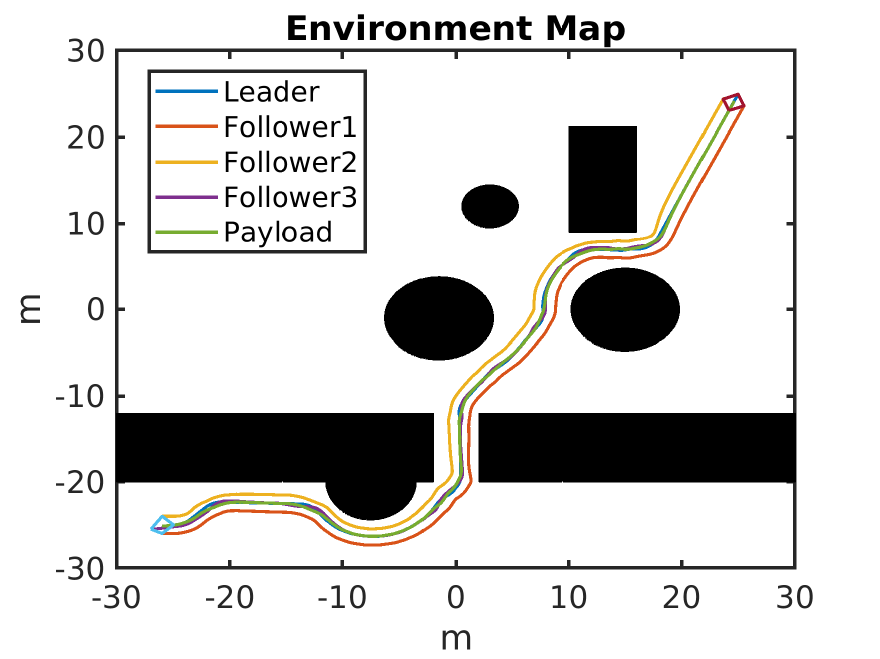}
\end{minipage}%
\begin{minipage}{0.5\linewidth}
\centering
\includegraphics[height=1.37in]{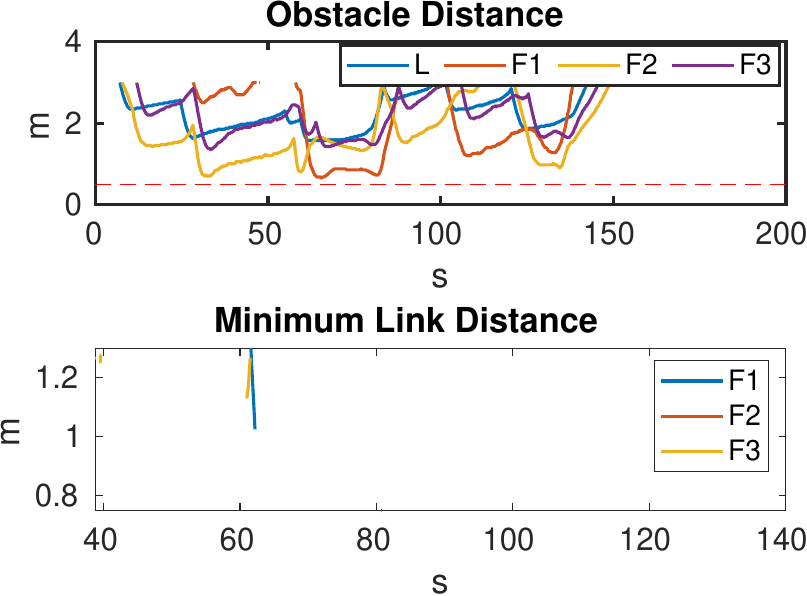}
\end{minipage}
\caption{Robot trajectories in the simulation environment, $\rho_{io}$ and $\rho_{il}$}
\label{Simulation_1}
\vspace{-0.3cm}
\end{figure}

Here, several simulations are provided to verify our proposed methods in MATLAB. As in Fig. \ref{Simulation_1}, the robot formation is required to transport the payload to $\bm{p}_g = [25,25]^T(m)$, while constraints in Equation (\ref{eq2a})-(\ref{eq2f}) should be maintained. As for scalable constraints, we select $\bar v_i^2+\vert \omega_i \vert \Vert {}^i\bm{p}_j \Vert \le 8$ and $\bar \omega_i + \Vert {}^i\bm{p}_j \Vert^2 \le 8$ for each follower $i$.               

In this simulation, the desired displacements of followers are set as ${}^1\boldsymbol{c}_l^d = [1,-1]^T(m)$, ${}^2\boldsymbol{c}_l^d = [1,1]^T(m)$ and ${}^3\boldsymbol{c}_l^d = [2,0]^T(m)$ with $R_f=1m$. The related parameters are $l=0.9m$, $d=1.5m$, $h_0=0.1m$, $h_{\min}=0.1m$, $R_{s}=0.3m$, $\rho_c=0.4m$ and $c_{\max}=10m$, which implies that $r_{+}=1.4$ and $r_{-}=0.3$. The upper bounds of leader velocities are $\tilde{v}_l=0.5m/s$ and $\tilde{\omega}_l=0.4rad/s$. As for camera parameters, $\delta_c=0.5rad$ and $k_c=0.2$. The tracking parameter $c_1$ is set as $0.4$, while $c_2=0.7$ and $c_3=0.4$. For the constraint preservation process, $k_v^o=0.2$, $\beta_t=0.5$, $d_o=0.5m$, $d_l=0.4m$, $\delta_r = 0.5m$, $\delta_o=0.8m$ and $\delta_l=0.8m$. In each figure, we use $L$ to represent the leader and $Fi$ to represent the follower $i$.

\begin{table}[!htbp]\label{table_1} 
\vspace{-0.35cm}
\centering
\caption{Our scheme versus other frameworks in Matlab}
\begin{tabular}{ccccccccccc}
\toprule 
\multicolumn{3}{c}{\multirow{2}*{Method}}& \multicolumn{2}{c}{Trajectory generation}& &\multicolumn{2}{c}{Trajectory tracking}\\
\cmidrule(r){4-9} 
\multicolumn{3}{c}{}&Mean\,(ms)&Std\,(ms)&&Mean\,(ms)&Std\,(ms)&\\ 
\hline 
\multicolumn{3}{c}{Cascade QP\cite{O.K.2011}}& -& -&  &730.1& 130.5\\ 
\multicolumn{3}{c}{HQP\cite{D.K.2021}}& -& -&  &415.3& 80.1\\   
\multicolumn{3}{c}{MIQP\cite{D.M.2011}}&31261.2&2514.6& &-&-\\
\multicolumn{3}{c}{dec\_iSCP\cite{Y.C.2015}}&10524.4&975.1& &-&-\\
\multicolumn{3}{c}{Ours}&0.091&0.013& &0.17&0.022\\
\bottomrule 
\end{tabular}
\vspace{-0.3cm}
\end{table}

\begin{figure} [htbp]
\vspace{-0.35cm}
\begin{minipage}{0.5\linewidth}
\centering
\includegraphics[height=1.2in]{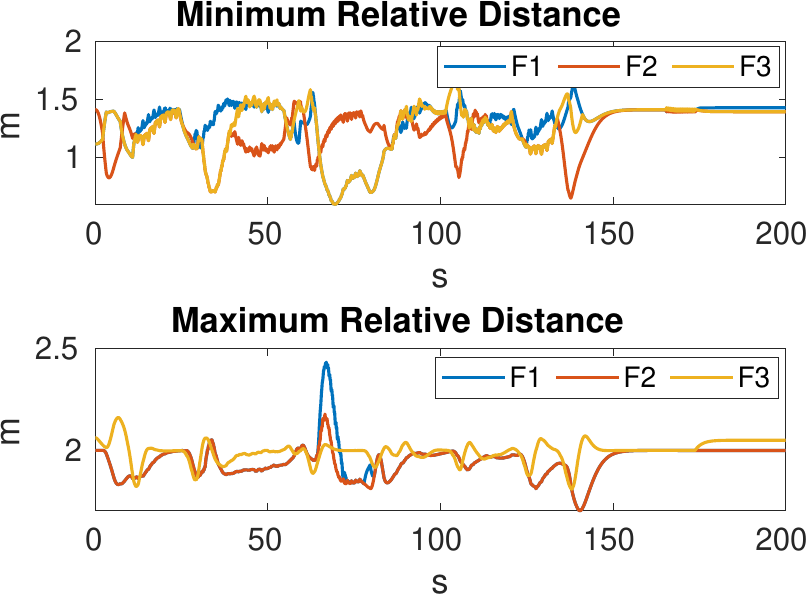}
\end{minipage}%
\begin{minipage}{0.5\linewidth}
\centering
\includegraphics[height=1.2in]{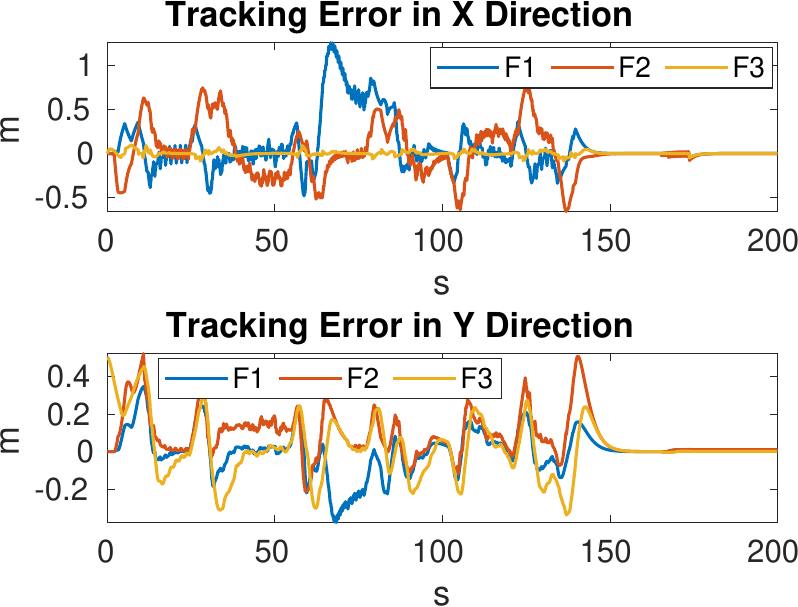}
\end{minipage}
\caption{The extreme values of $\Vert {}^i\bm{p}_j \Vert$ and tracking errors ${}^ix_e$, ${}^iy_e$}
\label{Simulation_2}
\vspace{-0.35cm}
\end{figure}

The tracking errors in $x$ and $y$ directions are depicted in Fig. \ref{Simulation_1} and the angle error is shown in Fig. \ref{Simulation_2} The trajectory tracking errors always converge to $0$ except instants when constraints tend to be destroyed. The obstacle distance $\rho_{io}$ is always kept larger than $d_o$ as in Fig. \ref{Simulation_1}, which implies that obstacle avoidance is achieved. Besides, the relative distances between robots are listed in Fig. \ref{Simulation_2}. Here, we define the minimum relative distance for the follower $i$ as $\rho_{ir}^{\min}=\min_{j\in \mathcal{N}_i}\{\Vert {}^i\bm{p}_j \Vert\}$ and the maximum relative distance as $\rho_{ir}^{\max}=\max_{j\in \mathcal{N}_i}\{\Vert {}^i\bm{p}_j \Vert\}$. For each follower, $\rho_{ir}^{\min}$ should be larger than $\sqrt{2}r_{-}=0.42m$ and $\rho_{ir}^{\max}$ less than $2r_{+}=2.8m$ under above constraints. It is found that all constraints for relative distances are satisfied.     

\begin{figure} [htbp]
\vspace{-0.3cm}
\begin{minipage}{0.5\linewidth}
\centering
\includegraphics[height=1.2in]{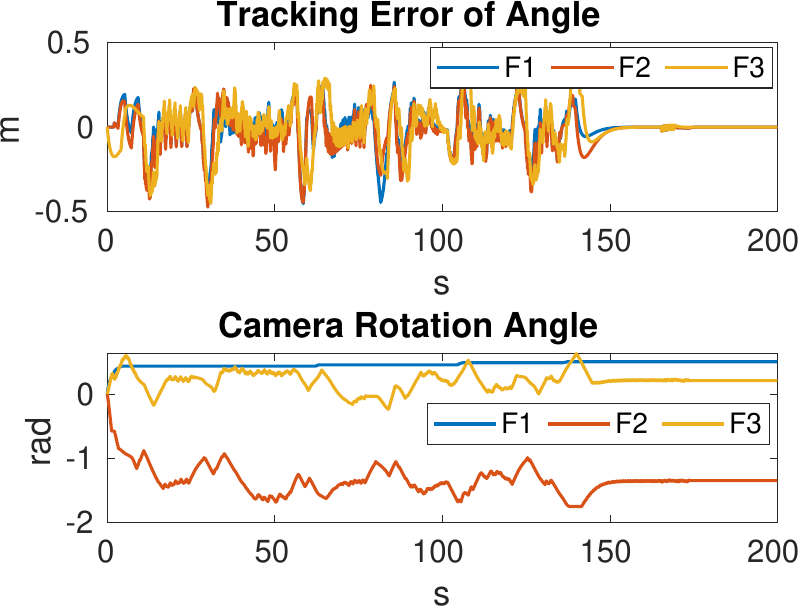}
\end{minipage}%
\begin{minipage}{0.5\linewidth}
\centering
\includegraphics[height=1.2in]{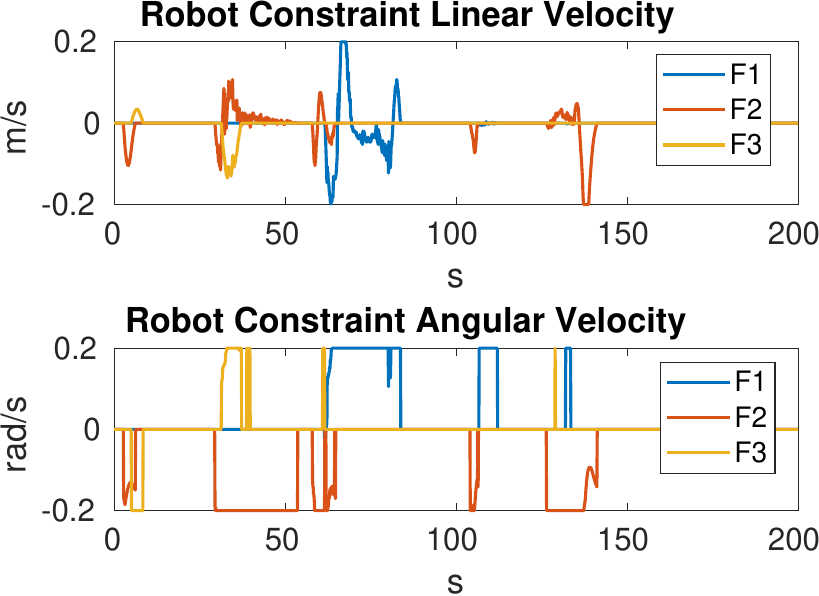}
\end{minipage}
\caption{The constraint velocities $v_i^o$ and $\omega_i^o$, ${}^i\theta_e$ and $\delta_c$}
\label{Simulation_3}
\vspace{-0.45cm}
\end{figure}

Another constraints are the link distance constraint in Equation (\ref{eq2d}) and velocity constraint. The link distance $\rho_{il}$ are maintained as in Fig. \ref{Simulation_1}. Here, the upper bounds of velocities are $\bar v = 1.8m/s$ and $\bar \omega = 1.8rad/s$. Based on above parameters, the constraint velocities should satisfy $\vert v_i^o\vert \le 0.2m/s$ and $\vert \omega_i^o\vert \le 0.2rad/s$ as in Fig. \ref{Simulation_3}. Furthermore, the rotation angles of cameras for leader tracking are shown beside the constraint velocities. In general, the system can finally transport the payload to the goal and preserve related constraints.       


Besides, we compare ours with other trajectory generation and trajectory tracking methods on our Acer Predator laptop with i7-12700H CPU as in Table \uppercase\expandafter{\romannumeral1}, where we randomly select $10$ execution results of each method to calculate the mean and standard deviation. Since most trajectory generation algorithms are done offline based on global maps, we record the time spent by them in offline planning. Compared to the tens of seconds they spend offline, our robots plan in real time and spend on average less than 0.1 ms per step. As for trajectory tracking under constraints, other methods like Cascade QP and HQP consumes a few hundred milliseconds per step, while our scheme only takes about 0.2 ms in a distributed configuration. Here, when calculating the consuming time of our method, we first average the consuming time of all robots as $t_a$. Then, the results are calculated by taking $t_a$ at different moments, which also applies to the experiment part.

\subsection{Robot Formation Experiments}
\begin{figure} [htbp]
\vspace{-0.35cm}
\begin{minipage}{0.5\linewidth}
\centering
\includegraphics[width=1.85in]{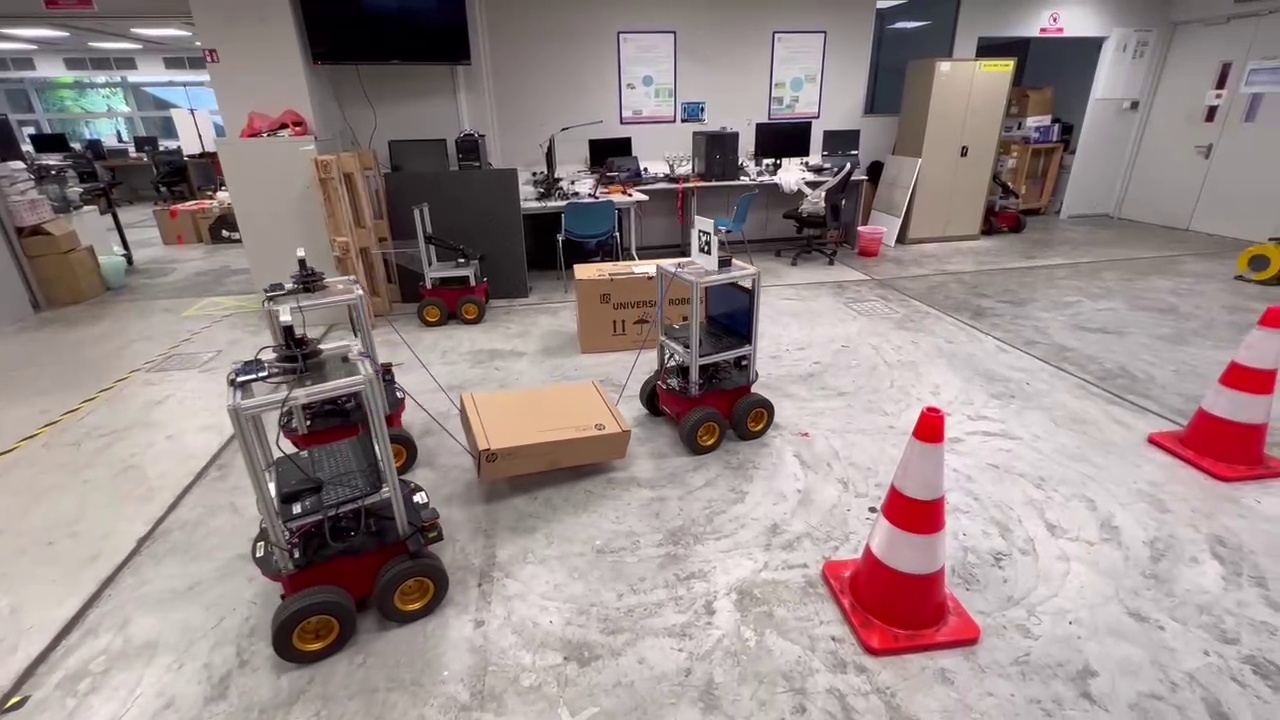}
\end{minipage}%
\begin{minipage}{0.54\linewidth}
\centering
\includegraphics[width=1.38in]{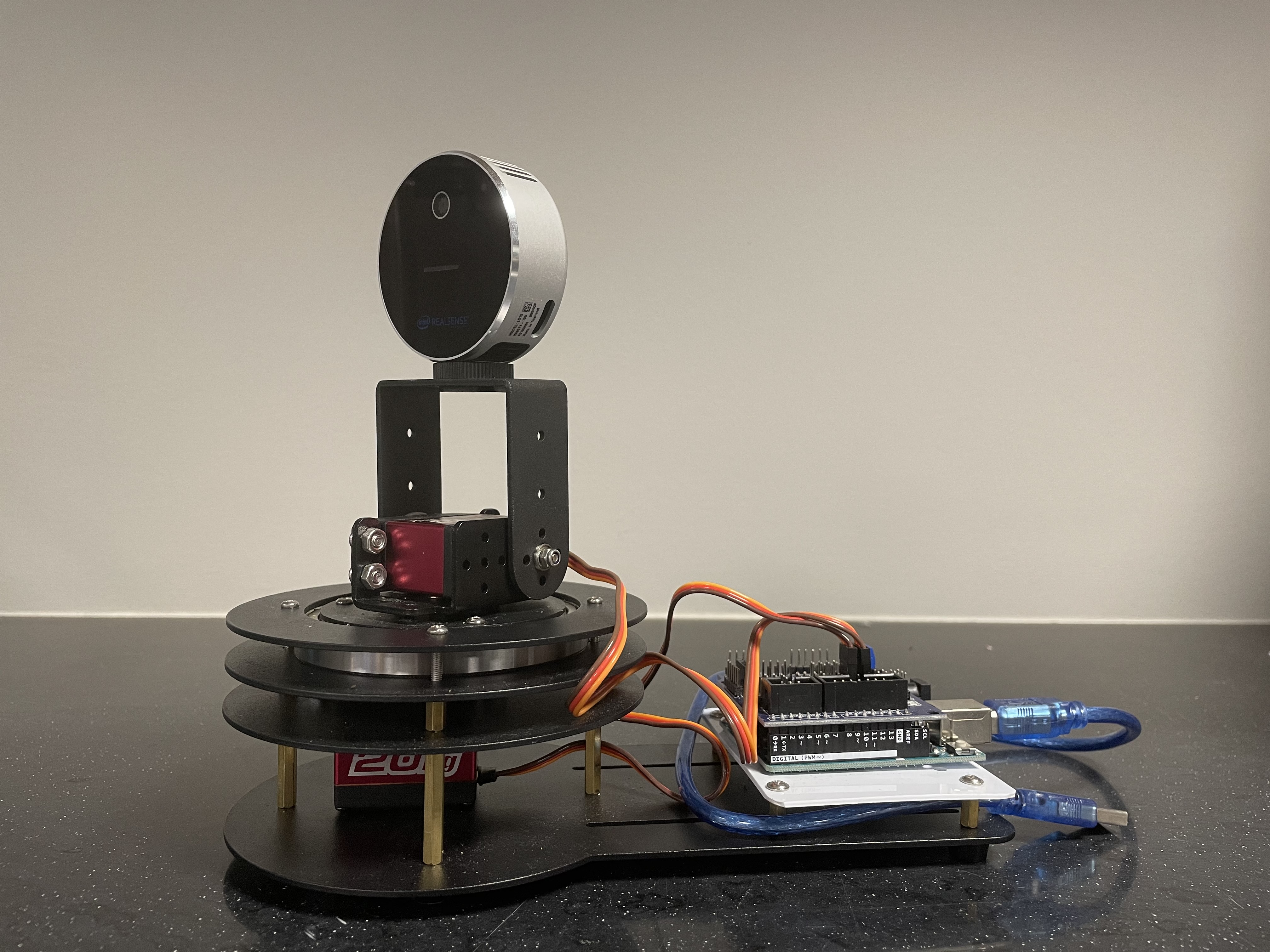}
\end{minipage}
\caption{Actual cooperative transportation system and the rotating camera}
\label{experiment}
\vspace{-0.35cm}
\end{figure}

In this part, the experiment is performed by three nonholonomic mobile robots to verify our strategies. As in Fig. \ref{experiment}, these robots are divided into one leader and two followers. The robot formation is required to transport the suspended payload to the goal $\boldsymbol{p}_g=[5.5, 2]^T(m)$, with constraints in Equation (\ref{eq2a})-(\ref{eq2f}) preserved. Besides, the video of simulation and experiment can be found via the link \href{https://youtu.be/qAHkj-s8STE}{https://youtu.be/qAHkj-s8STE}.    
%

\begin{figure} [htbp]
\begin{minipage}{0.5\linewidth}
\centering
\includegraphics[width=1.6in,height=1.3in]{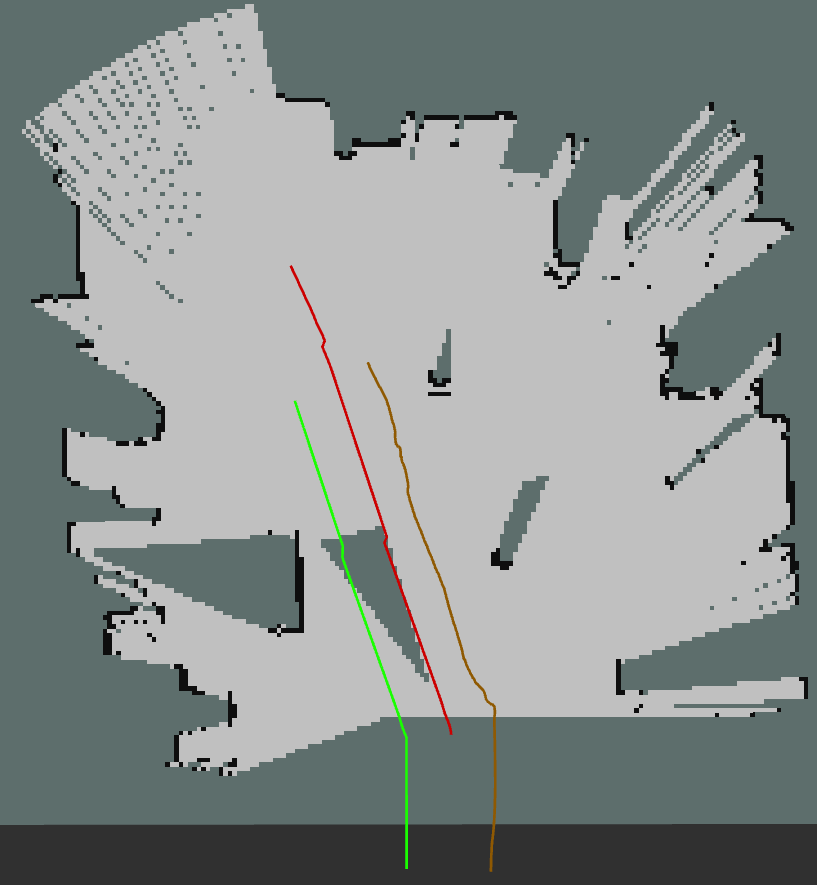}
\end{minipage}%
\begin{minipage}{0.5\linewidth}
\centering
\includegraphics[width=1.8in,height=1.4in]{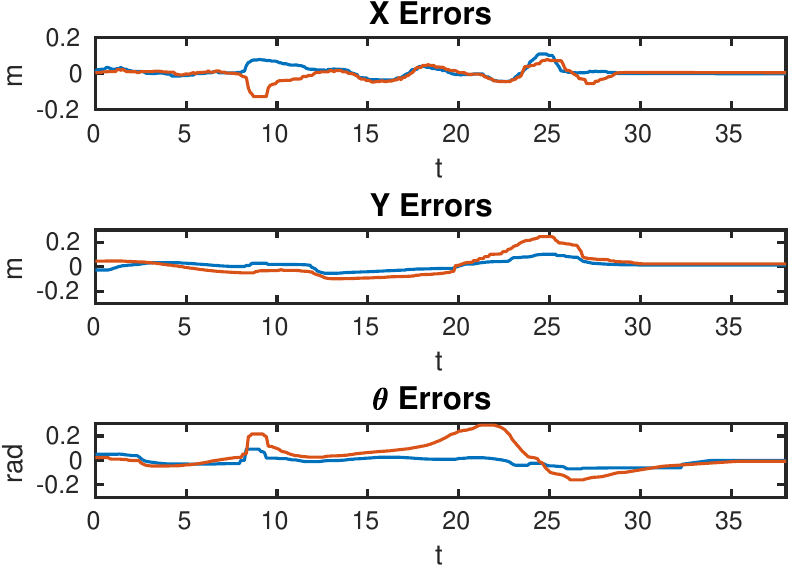}
\end{minipage}
\caption{Robot trajectories in RVIZ, ${}^i{x}_e$, ${}^i{y}_e $ and ${}^i{\theta}_e $}
\label{experiment_1}
\vspace{-0.6cm}
\end{figure}

The desired trajectory tracking displacements for followers are ${}^1\boldsymbol{c}_l^d =[1.4,-0.5]^T (m)$ and ${}^2\boldsymbol{c}_l^d =[1.4,0.5]^T (m)$ with $R_f \approx 0.8m$. The robot trajectories are illustrated in the left of Fig. \ref{experiment_1}, where the red line corresponds to the leader, the green line and brown line are for Follower 1 and 2. The trajectory tracking errors for followers are provided in the right, which converge to zero finally. In the following figures, the blue line records data about Follower 1, the red line is for Follower 2.                      


\begin{figure} [htbp]
\vspace{-0.3cm}
\begin{minipage}{0.5\linewidth}
\centering
\includegraphics[width=1.75in]{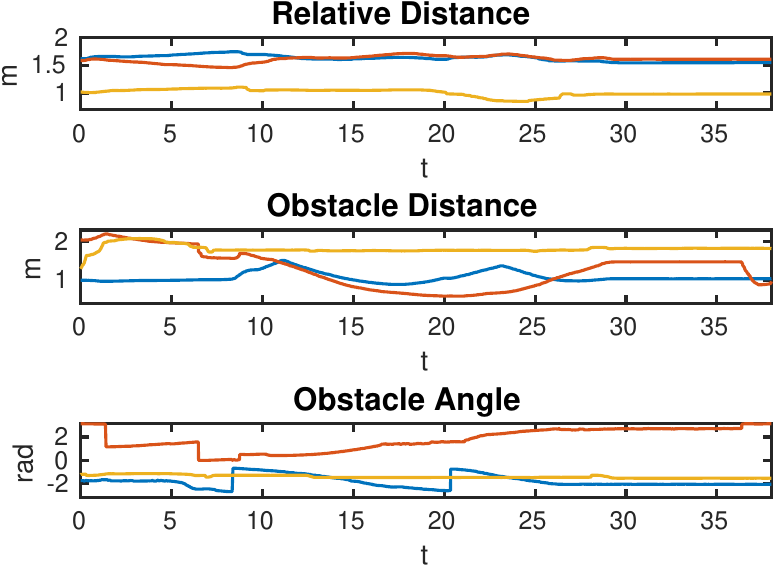}
\end{minipage}%
\begin{minipage}{0.5\linewidth}
\centering
\includegraphics[width=1.75in]{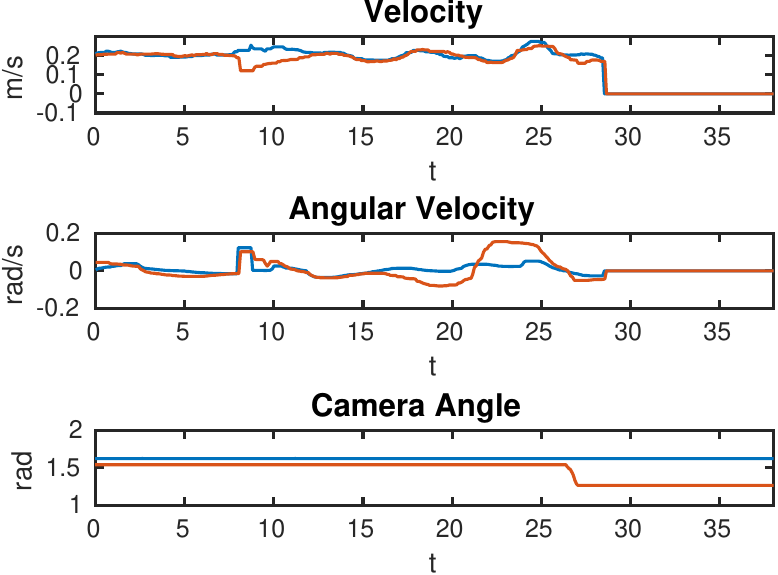}
\end{minipage}
\caption{$\Vert {}^i\bm{p}_j \Vert$, obstacle measurements, robot velocities and camera angles, where the yellow line describes $\Vert {}^1\bm{p}_2 \Vert$ in the first figure and the leader data in the following two.}
\label{experiment_2}
\vspace{-0.3cm}
\end{figure}

The relative distance $\Vert {}^i\bm{p}_j \Vert$, the obstacle distance $\rho_{io}$ and angle $\alpha_{io}$ are shown in Fig. \ref{experiment_2}. Here, $c_{\max}=5m$, $l=0.7m$, $d=0.83m$ approximately, $\rho_=0.5m$, $\rho_o=0.3m$, $\delta_r = 0.3m$, $\delta_o = \delta_l = 0.3m$, $R_s=0.36m$, $r_{+}=1.325$, $r_{-}=0.65$, $h_0=0.1m$ and $h_{\min}=0.05m$. Thus, the lower bound of distance between two followers is $\max\{\rho_c,r_{-}\}=0.65m$, while the upper bound is $\min\{c_{\max},r_{+}\}=1.325m$. For each leader-follower pair, the bounds are $0.97m$ and $1.96m$. As shown, the constraints of relative distance and obstacle avoidance are preserved. Besides, $\vert \alpha_{io} \vert$ changes towards $\frac{\pi}{2}$, which obeys the former analysis. The results of $\rho_{il}$ are omitted here, because they are nearly the same as $\rho_{io}$ in our lab environment.

As for velocities, the leader velocity is bounded by $\tilde{v}_l=0.2m/s$ and $\tilde{\omega}_l=0.3rad/s$, while the follower velocity is constrained within $\bar v=1.2m/s$ and $\bar \omega = \frac{5\pi}{6}rad/s$ according to robot datasheet. Here, $c_1=0.2$, $c_2=0.4$, $c_3=0.4$, $k_{v}^o=0.1$, $k_{\omega}^o=0.3$. As in Fig. \ref{experiment_2}, velocities of two followers are limited within robot maximum velocities. Furthermore, $\delta_c=0.4rad$ and $k_c=1$. As illustrated, the follower camera is rotated actively to catch the leader. From the experiment results, we know that the payload is transported to the goal, with all related constraints preserved. Furthermore, the average time for each update in our algorithm is only $0.0087 \pm 0.0011$ ms on C++ platform, which is quite time-efficient compared with HQP methods taking at least $4$ ms per update.

\section{Conclusion}
This paper proposes a new cooperative transportation scheme for nonholonomic mobile robots using a leader-follower approach. Unlike most existing methods, it combines low computational complexity with scalable constraints, is distributed, and operates without prior environment maps. However, it still depends on inter-robot communication and does not consider payload dynamics or force balance. Future work will address these limitations and explore more cost-effective solutions.

\begin{IEEEbiography}[{\includegraphics[width=1in,height=1.25in,clip,keepaspectratio]{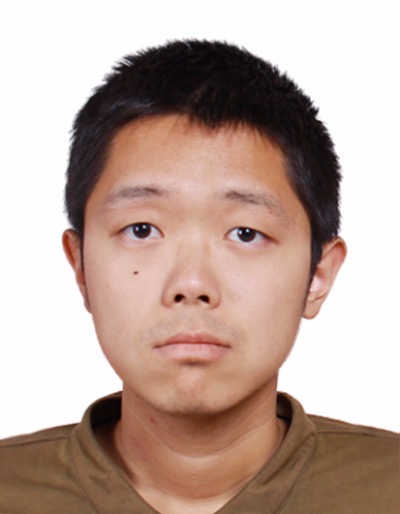}}]
{Renhe Guan} received the Bachelor of Engineering (B.E.) degree in Automation from Harbin Institute of Technology, Harbin, China, in 2018. He completed the Doctor of Philosophy (Ph.D.) degree in Electrical and Electronic Engineering at Nanyang Technological University, Singapore, in 2024. He is currently with Harbin Institute of Technology Shenzhen, Shenzhen,
China. His research interests include advanced multi-robot coordination strategies, cooperative transportation mechanisms and vision-based servo control.
\end{IEEEbiography}


\begin{IEEEbiography}[{\includegraphics[width=1in,height=1.25in,clip,keepaspectratio]{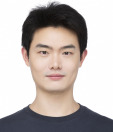}}]
{Tao Liu} received his B.Eng. degree from the University of Science and Technology of China in 2014 and Ph.D. degree from The Chinese University of Hong Kong (CUHK) in 2018. From 2019 to 2022, he worked as a Postdoctoral Fellow with the Department of Mechanical and Automation Engineering, CUHK, supported by the Impact Postdoctoral Fellowship of CUHK and the RGC Postdoctoral Fellowship of the Hong Kong Research Grants Council. He is now an Associate Professor with the College of Engineering, Southern University of Science and Technology, Shenzhen, China.
Dr. Liu is a recipient of the 2019 CAA Excellent Doctoral Dissertation Award. His research interests include output regulation, nonlinear control, cooperative control, and their applications to robotic systems.
\end{IEEEbiography}

\begin{IEEEbiography}[{\includegraphics[width=1in,height=1.25in,clip,keepaspectratio]{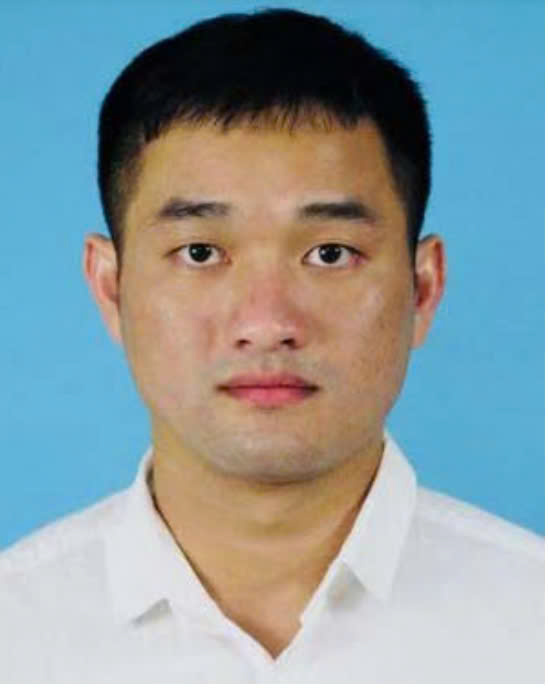}}]
{Yan Wang} received the B.E. degree in automation and the Ph.D. degree in control science and
engineering from the University of Science and
Technology of China, Hefei, China, in 2014 and
2019, respectively.

From 2019 to 2021, he was a Research Fellow with Nanyang Technological University, Singapore. From 2021 to January 2023, he was affiliated with The Hong Kong Polytechnic University, Hong Kong, and The Chinese University of
Hong Kong, Hong Kong. He is currently an Associate Professor with the School of Mechanical Electrical Engineering
and Automation, Harbin Institute of Technology Shenzhen, Shenzhen,
China. His research interests include optimal control/estimation of interconnected systems, connected vehicle system control and optimization,
and AGV schedule in flexible manufacturing systems.
\end{IEEEbiography}

%
%
%
%
%




\end{document}